# Development and Testing of a Novel Large Language Model-Based Clinical Decision Support Systems for Medication Safety in 12 Clinical Specialties

Jasmine Chiat Ling Ong1,2*
Jin Liyuan2,3,4*
Kabilan Elangovan3,4
Gilbert Yong San Lim3,4
Daniel Yan Zheng Lim5,6
Gerald Gui Ren Sng6,7
Ke Yu He8
Joshua Yi Min Tung6,9
Ryan Jian Zhong1
Christopher Ming Yao Koh1
Keane Zhi Hao Lee1
Xiang Chen1
Jack Kian Ch'ng10
Aung Than11
Ken Junyang Goh12
Daniel Shu Wei Ting2,3,4+

1. Division of Pharmacy, Singapore General Hospital, Singapore
2. Duke-NUS Medical School, Singapore, Singapore
3. Singapore National Eye Centre, Singapore Eye Research Institute, Singapore, Singapore
4. Singapore Health Services, Artificial Intelligence Office, Singapore
5. Department of Gastroenterology, Singapore General Hospital, Singapore
6. Data Science and Artificial Intelligence Lab, Singapore General Hospital, Singapore
7. Department of Anesthesiology, Singapore General Hospital, Singapore, Singapore
8. Department of Endocrinology, Singapore General Hospital, Singapore
9. Department of Urology, Singapore General Hospital, Singapore
10. Department of Vascular Surgery, Singapore General Hospital, Singapore
11. Department of Internal Medicine, Singapore General Hospital, Singapore
12. Department of Respiratory and Critical Care Medicine, Singapore General Hospital, Singapore

*Contributed Equally
+Corresponding Author

**Corresponding author:**
A/Prof Daniel Ting MD (1st Hons) PhD
Associate Professor, Duke-NUS Medical School
Director, AI Office, Singapore Health Service
Address: The Academia, 20 College Road, Level 6 Discovery Tower, Singapore, 169856

**Abstract**

**Importance**: We introduce a novel Retrieval Augmented Generation (RAG)-Large Language Model (LLM) framework as a Clinical Decision Support Systems (CDSS) to support safe medication prescription, a critical aspect of patient safety. This overcomes existing challenges of irrelevancy of alerts in rules-based CDSS in provision of prescribing error alerts that is relevant to the patient's context and institutional medication use guides.

**Objective**: To evaluate the efficacy of LLM-based CDSS in correctly identifying medication errors in different patient case vignettes from diverse medical and surgical sub-disciplines, against a human expert panel derived ground truth. We compared performance for under 2 different CDSS practical healthcare integration modalities: LLM-based CDSS alone (fully autonomous mode) vs junior pharmacist + LLM-based CDSS (co-pilot, assistive mode).

**Design, Setting, and Participants**: Utilizing a RAG model with state-of-the-art medically-related LLMs (GPT-4, Gemini Pro 1.0 and Med-PaLM 2), this study used 61 prescribing error scenarios embedded into 23 complex clinical vignettes across 12 different medical and surgical specialties. A multidisciplinary expert panel assessed these cases for Drug-Related Problems (DRPs) using the PCNE classification and graded severity / potential for harm using revised NCC MERP medication error index. We compared.

**Main Outcomes and Measures**: This study compares the performance of an LLM-based CDSS in identifying DRPs. Key metrics include accuracy, precision, recall, and F1 scores. We also compare the performance of LLM-CDSS alone and junior hospital pharmacists (less than 2 years post licensure) + LLM-CDSS (co-pilot, assistive mode) in the provision of recommendations to clinicians. In addition, we present comparative results from different LLMs: GPT-4, Gemini Pro 1.0 and Med-PaLM 2.

**Results**

RAG-LLM performed better compared to LLM alone. When employed in a co-pilot mode, accuracy, recall, and F1 scores were optimized, indicating effectiveness in identifying moderate to severe DRPs. The accuracy of DRP detection with RAG-LLM improved in several categories but at the expense of lower precision.

**Conclusions**

This study established that a RAG-LLM based CDSS significantly boosts the accuracy of medication error identification when used alongside junior pharmacists (co-pilot), with notable improvements in detecting severe DRPs. This study also illuminates the comparative performance of current state-of-the-art LLMs in RAG-based CDSS systems.

**Introduction**

Medical errors remain a formidable challenge for healthcare institutions all around the world and is the third leading cause of mortality in the United States. Medication-related errors account for 5% to 41% of US hospital admissions every year.[1,2] Medication errors can potentially result in prolonged hospitalization stay, elevated risk for morbidity and mortality as well as an increase in healthcare spending.[3] This translates to high economic burden of medication errors, amounting up to USD$40 billion in the United States and £750 million in England per year.[3,4] In an acute care setting, medication errors can occur at any stage of the medication use process: medication prescribing, dispensing, administration and patient monitoring. A vast majority of errors occur at the prescribing stage, accounting for 70% of errors that result in adverse patient events.[5] Halting errors at this stage is critical in preventing perpetuation of the error downstream and eventually reaching the patient.

Clinical Decision Support Systems (CDSS) have become a cornerstone of modern healthcare systems as a direct aid to clinical decision making. A CDSS is intended to improve healthcare delivery by enhancing medical decisions with targeted clinical knowledge, patient information, and other health information.[6] A key function of CDSS is in reducing the incidence of prescribing or medication errors and adverse events through integration with electronic health records and computerized provider order entry (CPOE) systems.[7] In particular categories of prescribing errors such as drug-drug interactions, CDSS have been shown to be effective in reducing the incidence of errors.[8] However, a vast majority of current systems are rules-based, resulting in the generation of voluminous and clinically irrelevant alerts to users.[9] Over time, safety alerts are ignored and overwritten by physicians in as high as 95%, this phenomenon of 'alert fatigue' leads to ignoring of these alerts, posing a barrier to effective adoption and utilization of CDSS systems.[10-12] There is a need to re-invent delivery of prompts by CDSS through greater personalization, intuitive and lesser emphasis on disruptive alerts.

The growing capabilities of large language models (LLMs) in medical tasks are becoming more apparent. LLMs are advancing in handling such tasks, particularly those that do not require extensive specialized expertise.[13,14] This includes simplifying administrative duties like composing medical letters, creating summaries upon patient discharge using information from electronic health records (EHR) to semi-autonomous decision-making support for managing operating theatres.[15,16] Finally, LLM-powered healthcare chatbots is capable of providing patients and health professionals with highly professional-sounding, accurate and personalized responses to medical queries.[17,18] Healthcare systems are facing critical shortage of healthcare professionals compounding by a mounting issue of healthcare professional burnout.[19-21] LLM-powered solutions are well poised to improve operational efficiency and standards of patient care when trained with the right data, robust clinical evaluation with a deployment strategy armed with appropriate safety measures.

Prior published studies have developed LLM-based tools to support various clinical applications and domains.[22-24] The use of LLMs as an innovative substitute to current rules-based CDSS has not been described. While hundreds of pretrained LLMs have published since last year, few LLMs are trained specifically on medical knowledge or reported their clinical performance. In this study, GPT-4.0, Gemini pro 1.0, and Med-PaLM 2 were chosen for their reported close to expert performance in various medical tasks.[25-28]

Herein, we propose a new model of CDSS leveraging upon LLM grounded with contextual knowledge on medications to accurately identify prescribing errors and provide evidence-based recommendations to healthcare professionals. In this study, we seek to address the following question: Can LLM-based CDSS identify prescribing errors and provide accurate, clinically acceptable recommendations across different medical disciplines? The results will reveal the potential for integration of LLMs in healthcare, potentially leading to substantial improvements in patient safety and quality of care.

**Methods**

We followed the Standards for Reporting of Diagnostic Accuracy (STARD) reporting guideline for this study on the development of LLM-based CDSS for medication prescribing error identification. An overview of the evaluation workflow is shown in Figure 1. Ethics review board was not required because no identifiable patient data were used.

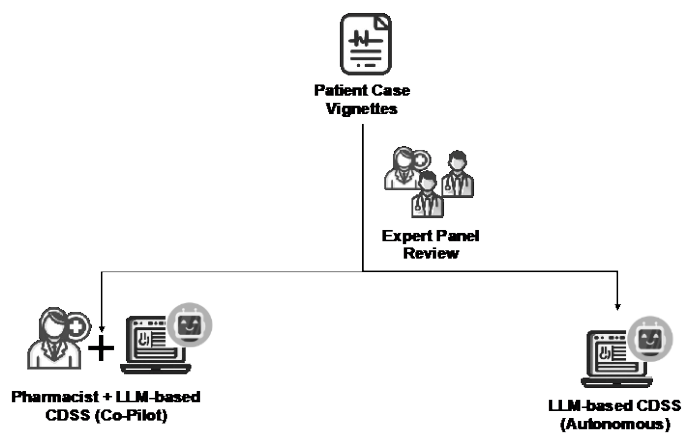

*Figure 1: Overview of evaluation workflow*

**Development of Prescribing Error Scenarios**

We created a total of 61 different simulated prescribing error scenarios based on 23 case vignettes modelled after complex clinical cases. Prescribing error scenarios were adapted from the local institution pharmacy intervention and error reporting databases to maintain realism. The vignettes covered clinical scenarios from 12 different medical or surgical subspecialties (Cardiology, Endocrinology, General Medicine, Ophthalmology, Gastroenterology, General Surgery, Urology, Vascular Surgery, Infectious Disease, Respiratory Medicine, Oncology, Colorectal Surgery), with some cases involving more than one discipline. Each case vignette consisted of a patient clinical note and medication prescription. Prescribing errors in the clinical scenarios were designed to be reflective of commonly encountered drug-related problems (DRP). We present a sample of one case vignette in Figure 2. A detailed description of all case vignettes is found in the supplement. The Anatomical

Therapeutic Chemical (ATC) category of medications prescribed in each clinical scenario is also presented in the supplement.

```
CVM Inpatient Daily Ward Round
General Information:
Admission Date 11-Sep-2024 14:24:49 Post Admission Day 2

Clinical Notes:
Latest Vital Signs: 12-Sep-2024 07:00:00
    12/09/2024 07:00:00
    Pain Score: 0
    BP (NIBP) (mmHg): 103/83 (101-144/68-118), HR (beats/min): 65 (55-99)
    RR (breaths/min): 12 (12-21), SPO2 (%): 100 (95-100), O2 Therapy (L/min): NP 2 (2-3)
    Hypocount (from 11/09/2024 06:00:00 to 12/09/2024 07:18:37):
    15.8(H) < 20.4(H) < 27.5(HH) < 21.0(H) < 5.9(N)
    Ht: 182 cm (11-Sep-2024 15:18:00), Wt: 99.1 kg (11-Sep-2024 16:57:00)
    BMI: 29.9, BSA: 2.24 m2
    12/09/2024 04:00:00
    T (deg.C): 36.6, Tmax (deg.C): 36.6 (12/09/2024 04:00:00)

Latest I/O from 11/09/2024 06:01 to 12/09/2024 06:00
Intake: 2100    Output: 3750    Net: -1650
Intake: 2100
- Diet Fluid Volume: 100
- IV: 2000
Output: 3750
- Urine Output: 3750

Lab Values:
    12/09/2024 06:15
    Hb:     11.3 [12.0 – 16.0 G/DL]
    TW:     8.64 [4.0 – 10.0 x 10(9)/L]
    Plt:    250 [140 – 440 x 10(9)/L]
    SCr:    100 [37 – 75 UMOL/L]
    INR: 1.2

Surgical operations:
right LL deep vein thrombolysis and creation of SFA-LSV arteriovenous fistual on 15-May-2024

Clinical Notes:
61 / M / Malay
Allergic to mefenamic acid and salicylate, claims rashes and facial swelling
Lives with wife and children
ADL independent, Community ambulant

PMHx:
1. Poorly controlled DM
2. HTN, HLD
3. CKD Stage 2
4. Necrotising fasciitis of R LL
  - s/p R BKA 12/2/20
5. Left foot OM
  - s/p 2nd and 3rd ray amputation Dec 2019
6. Large right medial thigh abscess
  - s/p I&D 6/3/24
7. Recurrent DVT and PE (on long-term warfarin) - f/u Haem and VAS
  - 2018: L LL DVT
  - 2020: Extensive left arm DVT s/p R LL thrombolysis and IVC filter insertion
  - s/p R LL venous thrombectomy with stenting 28/4/24
  - Possible right interlobar pulmonary embolism (April 2024)

HOPC:
Chest pain since Friday
Started at rest
Central pressing with dyspnoea and diaphoresis
No radiation
Stuttering since then
Worse with exertion
Worse since 5am today hence presented
Still has ongoing mild discomfort
No fever or intercurrent illness
No bleeding history
Normally takes warfarin at night - has not taken for today
All his brothers had CABG at his age
Single NSAID (mefenamic acid) allergy - rashes
Progress in emergency department:
Bedside echo: LV impaired systolic function (also does look dilated), RV normal size and function, EF < 45%
ECG: evolved anterior MI (anterior Q waves new since May 2024)
Cath lab activated

O/E:
Vitals stable, afebrile
SpO2 95-100% on 2LNP
Poor H/C control
Telemetry ON: NSR
JVP not elevated
H S1 S2 NIL murmurs
L Clear
Abdomen SNT, BS+

Issues and progress:
1) Evolved anterior MI with ongoing chest discomfort
- Cath activated
- Findings:
- Co-dominant coronary system
- Severe proximal to mid LAD stenosis, involving the bifurcation with the dominant diagonal branch
- Diffusely diseased AV groove branch (continuation of the distal LCx)
- Anomalous origin of the RCA (high anterior). Diffuse moderate proximal to mid RCA narrowing
- Successful PCI to the LAD and implantation of overlapping drug eluting stent (Onyx) from mid to proximal LAD. Ostium of the diagonal branch preserved with modified jailed balloon technique.
- Post cath stable - to ICA
2) b/g of recurrent VTE on Warfarin
Hold off warfarin for now
Bridge with clexane
Trend Hb whilst on DAPT + anticoagulation
- Hb on 12/9 11.3 (stable)

Plan:
To GW with telemetry
Vitals as per ward protocol
CBG TDS + 10PM with SCSI cover
Low salt/low fat/DM diet
Heart failure for medical therapy in the interim
Aspirin for 1/12
Clopidogrel for at least 12 months

Allergies:
Mefenamic Acid. Facial Swelling.
Salicylate. Facial Swelling
```

| Medications Prescribed | Status |
|---|---|
| Sodium Chloride 0.9% InFUSion, IV Intermittent 2,000 mL, Once, Infuse Over 16 hour, 125 mL/hr | Active |
| Enoxaparin Sodium Injection, Sub-Cutaneous 60 mg, BD | Active |
| ACTRAPID [Insulin Soluble] Injection, Sub-Cutaneous 4 unit, Once | Active |
| LANTUS [Insulin Glargine] Solostar, Sub-Cutaneous 24 unit, OM | Active |
| NovoRAPID [Insulin Aspart] Flexpen, Sub-Cutaneous 8 unit, TDS (Pre-meal) | Active |
| Aspirin Tablet, PO 100 mg, OM | Active |
| Clopidogrel Tablet, PO 75 mg, OM | Active |
| OMEprazole Capsule, PO 20 mg, OM | Active |
| Glyceryl Trinitrate Tablet, Sub-Lingual 0.5 mg, Use as directed PRN Chest Pain | Active |
| Linagliptin Tablet, PO 5 mg, OM | Active |
| Bisoprolol Fumarate Tablet, PO 2.5 mg, OM | Active |
| Perindopril Erbumine [Tert-butylamine] Tablet, PO 2 mg, OM | Active |
| Neurobion Tablet [Vit B1 100mg, B6 200mg, B12 200mcg], PO 1 tablet, OM | Active |

Figure 2: Sample of Case Vignette. Abbreviations (CVM: cardiovascular Medicine, BP: blood pressure, RR: respiratory rate, Ht: height, BMI: body mass index, T: temperature, Hb: hemoglobin, TW: total white count, Plt: platelet count, SCr: serum creatinine, INR: international normalized ratio, PMHx: past medical history, DM: diabetes mellitus, HTN: hypertension, HLD: hyperlipidemia, CKD: chronic kidney disease, BKA: below knee amputation, OM: osteomyelitis, I&D: incision and drainage, DVT: deep vein thrombosis, PE: pulmonary embolism, VAS: vascular, HOPC: history of presenting complain, CABG: coronary artery bypass grafting, NSAID (non-steroidal anti-inflammatory drug), LV: left ventricle, RV: right ventricle, EF: ejection fraction, MI: myocardial infarction, O/E: on examination, NSR: normal sinus rhythm, JVP: jugular venous pressure, SNT: soft non tender, BS: bowel sound, LAD: left anterior descending artery, AV: atrioventricular, LCx: left circumflex artery, ICA: intermediate care area, DAPT: dual anti-platelet therapy)

**Development of Reference Standard**

The reference standard was developed through manual grading of DRP categories and severity by a multi-disciplinary expert panel. The expert panel consisted of pharmacotherapy board certified pharmacists and physicians with > 10 years of clinical practice experience in tertiary hospitals. Every clinical scenario was graded by at least 1 pharmacist and 1 physician member of the panel. Any disagreements were resolved by a 3rd independent member. DRPs from each clinical scenario was categorized using the Pharmaceutical Care Network Europe (PCNE) classification V9.1 and ASHP statement of pharmaceutical care as a guide.[29,30] Categories including in error scenarios include: adverse drug reaction, allergy, drug-drug interactions, duplication of therapy, inappropriate choice of therapy, inappropriate dosage regimen, no indication and omission of drug therapy. The potential severity of these errors were graded according to the Harm Associated with Medication Error Classification (HAMEC) classification tool.[31]

**Development of LLM-based CDSS Tool**

*Development of RAG-LLM Tool*

We engineered 2 versions of LLM-based tool using a Retrieval-Augmented Generation (RAG) and GPT4.0 as the base model. Showcasing both a simple (version 1) and more complex design (version 2) of RAG system. The RAG-LLM framework merges the Retrieval-Augmented Generation (RAG) model with a Language Model (LLM), optimizing the transformation of specialized documents into embedding vectors. This process utilizes advanced pre-processing and embedding models, with a focus on similarity-based retrieval between query vectors and embedded vectors of targeted documents, including drug monographs and hospital drug-use protocols.

Version 1 of the RAG-LLM framework utilized Pinecone for vector storage and OpenAI's text-embedding-ada-002 for embedding, with pre-processing by the Langchain Python package. It operated with a chunk size of 1000 and a retrieval parameter 'k' of 5, balancing retrieval detail and computational efficiency. Version 2 advanced the framework by integrating the Llamaindex RAG framework, relying on auto-merging retrieval to provide contextualized search. This version introduced manual indexing of drug names for targeted pharmaceutical relevance. It employed HuggingFace's bge-small-en-v1.5 embedding model and adjusted to hierarchical chunking sizes of 2048, 512, and 123. The 'k' value was increased to 20, enhancing the breadth and depth of information retrieval.

The prompt was designed as a series of tasks. Each task required retrieval of knowledge for each medication on the prescription (e.g. Is this medication indicated for the patient?), before feeding into GPT-4 to generate a response. The response from each task is fed into a final GPT-4 model for summarization and recommendations. A diagrammatic view of the sequence is shown in Figure 3. All responses were generated and analysed in triplicates to account for reproducibility. The performance of both versions was compared and one version was selected as the final co-pilot.

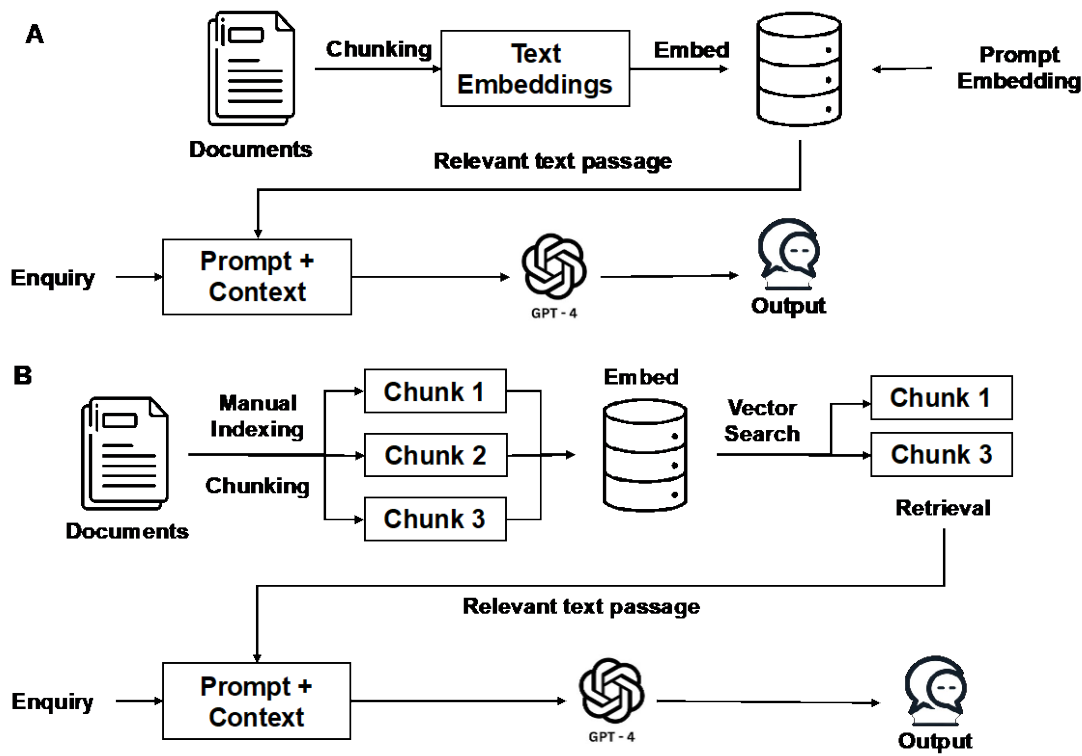

**Figure 3:** (A) Version 1 employing simplified RAG architecture (B) Version 2 employing advanced RAG architecture with auto-merging retrieval

*Knowledge Source*

Institutional medication use and dosing guidelines, medication monographs were used as sources of information. Each medication monograph was split into 4 separate sections according to the following categories of information: (1) Adverse drug reactions, cautions and contraindications, (2) ATC category and mechanism of action, (3) Drug-drug interactions, (4) Drug dosing and adjustments.

*LLM Prompt*

We first designed a general natural language prompting template and experimented with various prompting strategies to test the effect on model response. Various strategies that were tested included dynamic few-shot learning using 2 clinical scenarios separately created for in-context learning, chain of thought prompting and self-generated chain of thought prompting.[32] The final adapted prompt used is presented in Figure 4.

> **For the Junior Pharmacist**
> Assume the role of a clinical pharmacist. You are tasked to perform a medication chart review for a patient admitted to the department of <cardiology>. I will provide you with the patient's medication list, clinical note, and drug monographs as reference. Identify drug related problems specific to the patient's profile using this guide:
>
> - Medication Indications: Confirm that each medication has a clear indication and that current health conditions are being addressed with appropriate pharmacotherapy. [Drug monograph reference sections: "Pharmacologic Category", "Use: Labeled Indications", "Use: Off-Label: Adult", "Mechanism of Action"]
> - Dosing Verification: Check that the dosages of medications are within the recommended ranges and adjust if necessary, considering factors such as age, kidney function, and liver function [Drug monograph reference sections: "Dosing: Adult", "Dosing: Older Adult", "Dosing: Altered Kidney Function: Adult", "Dosing: Hepatic Impairment: Adult"]
> - Drug-Drug Interactions: Investigate potential interactions between current medications that could increase the risk of adverse effects or reduce therapeutic efficacy and warrants a change in therapy or monitoring tests. [Drug monograph reference sections: "Metabolism/Transport Effects", "Drug Interactions"]
> - Potential adverse drug reaction, contraindications and cautions, medication allergy [Drug monograph reference sections: "Special alerts", "ALERT: U.S. Boxed Warning", "Warnings/Precautions", "Contraindications", "Adverse Reactions", "Adverse Reactions (Significant): Considerations"]
> - Medication Omissions: Look for any conditions that are not being treated which should be, according to the patient's history and current clinical guidelines.
> - Any duplication in medication class or therapy
> - Patient-Specific Factors: Take into account patient-specific factors such as age, allergies, and preferences that may influence medication selection and management.
>
> Create a pharmacist recommendation note to address any identified drug related problem(s) in the following format: "situation, background, assessment, recommendation". Your plan should be clear and justified with specific recommendations for any changes to the medication regimen, including discontinuations, dose adjustments, or additions.

Figure 4: Final Adapted Prompt for LLM

**Generation of DRP Responses**

We generated GPT-4, version 1 RAG-LLM and version 2 RAG-LLM responses in triplicates, resulting in a total of 149, 161 and 246 DRP responses for assessment. Respective performance was used to select the best model to be used as comparator against human only and for adoption in co-pilot mode.

We randomly assigned the scenarios to 3 junior pharmacists (< 2 years of post-licensure practice experience) to independently identify any DRPs and generate a standard clinical note in standardized healthcare SBAR (situation, background, assessment, recommendation) format for each scenario. Each pharmacist had access to all institutional protocols and guidelines and was given a time limit of 1 hour to review and generate a set of responses for 7-8 clinical scenarios. Pharmacists were also given RAG-LLM generated responses as reference. During evaluation phase of different modes of deployment, a total of 52 and 39 DRP responses generated by co-pilot mode and autonomous mode respectively were assessed.

**Assessment of Accuracy of Responses**

We evaluated the accuracy of all DRP responses (see above section) using the human expert panel as the criterion standard. A DRP is considered correctly identified if the response mentions the specific medication name and provides a recommendation to perform an action or intervention to change or amend the medication order. If only the drug class is mentioned; or no action is suggested, the DRP is not considered accurately identified.

**Statistical Analysis**

The concordance of co-pilot and autonomous mode with human expert as benchmark was measured using accuracy, precision, recall, and F1 score as per a previous published study.[33] Accuracy is expressed as a percentage of correctly identified DRP against expert. Precision, which denotes the fraction of DRPs correctly identified amongst all suggested DRPs, was defined as precision = true positives / (true positives + false positives). Recall, or the fraction of all DRPs in the criterion standard correctly identified by LLMs, was defined as recall = true positives / (true positives + false negatives). The F1 score is the harmonic mean of precision and recall, and thus penalizes unbalanced precision and recall scores (ie, is higher when both precision and recall have similar values): F1 score = (2 × precision × recall) / (precision + recall). The higher any of the 3 scores, the better the response, with 1 being the maximum value for each score. Data analysis, calculation of precision, recall and F1 scores were done in R version 4.3.0 (R Project for Statistical Computing).

**Results**

Summary characteristics of error scenarios and case vignettes are presented in the supplement. Briefly, the case vignettes are representative of complex clinical case scenarios with multiple co-morbidities and problem lists. The expert panel determined that 29.5% of error scenarios were capable of causing serious harm, 50.8% capable of inflicting moderate harm while the remaining 19.2% were rated as capable of causing minor or no harm. The three most common DRPs in the case scenarios were: Inappropriate dosage regimen that arose due the need for dose, frequency or duration adjustment; adverse drug reactions requiring change in medication or reversal agents (this also includes prescribing medications that are contraindicated based on patient profile); and significant drug-drug interactions requiring change in medication or therapeutic drug monitoring. The median unique medications across ATC classes was 12 (IQR 5 – 16) per case vignette, suggesting that our constructed cases were complex and demonstrative of patients seen at a tertiary healthcare institution.

**Comparative Performance of GPT-4 and RAG-LLMs**

We first performed evaluation of results from GPT-4, version 1 and version 2 of RAG-LLM. We found that GPT-4 scored the lowest on measures of overall accuracy, recall and F1. While Version 2 RAG-LLM performed the best in terms of accuracy and recall, precision dropped drastically due to the high volumes of false positive responses generated. This trade-off is indicative of Version 2's sophisticated features like auto-merging retrieval and manual indexing, which, while enhancing recall and accuracy, compromise precision. Incorporating a more complex embedding model and larger 'k' values in version 2 may not have been optimally aligned with the dataset as expected thus impacting precision negatively. As such, we chose adopt version 1 RAG-LLM for subsequent evaluation as co-pilot in view of the highest F1 score and highest precision. This model presents with the best balance between achieving higher accuracy of identifying DRPs at the lowest alert burden (false positive rate).

|  | GPT-4 | | Version 1 RAG-LLM | | Version 2 RAG-LLM | |
| --- | --- | --- | --- | --- | --- | --- |
|  | Mean | SD | Mean | SD | Mean | SD |
| Precision | 0.368 | 0.046 | 0.407 | 0.045 | 0.317 | 0.020 |
| Recall | 0.295 | 0.033 | 0.355 | 0.009 | 0.426 | 0.028 |
| F1 | 0.325 | 0.013 | 0.379 | 0.022 | 0.364 | 0.023 |
| Accuracy (%) | 29.500 | 3.300 | 35.467 | 0.924 | 42.600 | 0.280 |

Table 1: Comparative performance of GPT-4, Version 1 and Version 2 of RAG-LLM. Poorest performance highlighted in red while best performance highlighted in green.

**Comparative performance of RAG-LLM and Co-pilot modes**

The co-pilot mode demonstrated the highest accuracy. Using RAG-LLM as a co-pilot doubled the relative accuracy of DRP identification as compared to RAG-LLM alone (54.1% Co-pilot vs 31.1% autonomous). LLM (without RAG) performed the poorest. Precision, Recall and F1 score was also highest in the co-pilot mode. In terms of DRP categories, co-pilot mode accuracy improved across most categories including adverse drug reaction, drug-drug interaction, duplication of therapy, inappropriate choice of therapy and omission of drug therapy. However, performance for co-pilot mode declined in the categories of inappropriate dosage regimen and no indication when pharmacists were unblinded to RAG-LLM response.

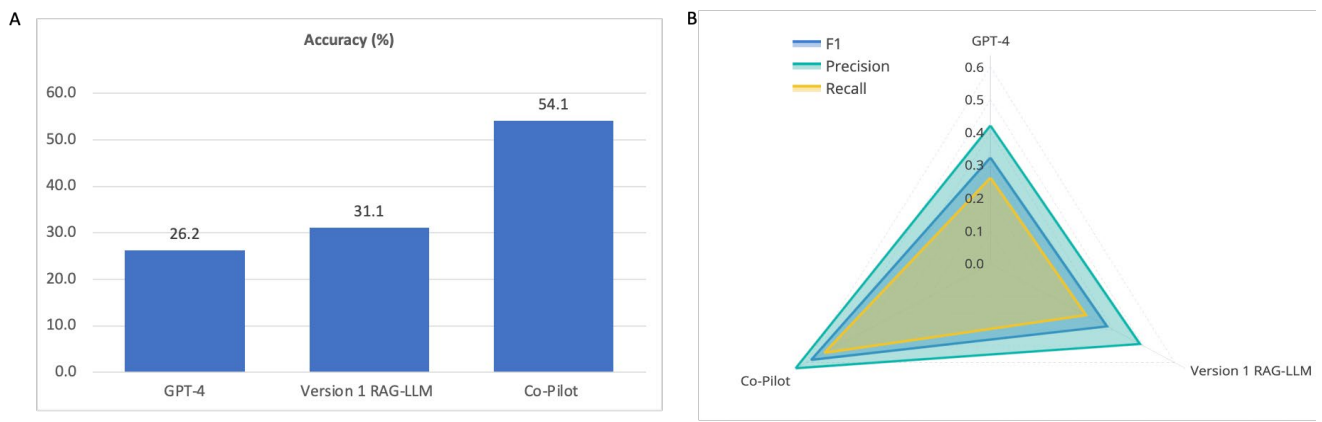

Figure 5: (A) Chart showing comparative accuracy of different modes. (B) Spider diagram showing relative precision, recall and F1 scores of different modes.

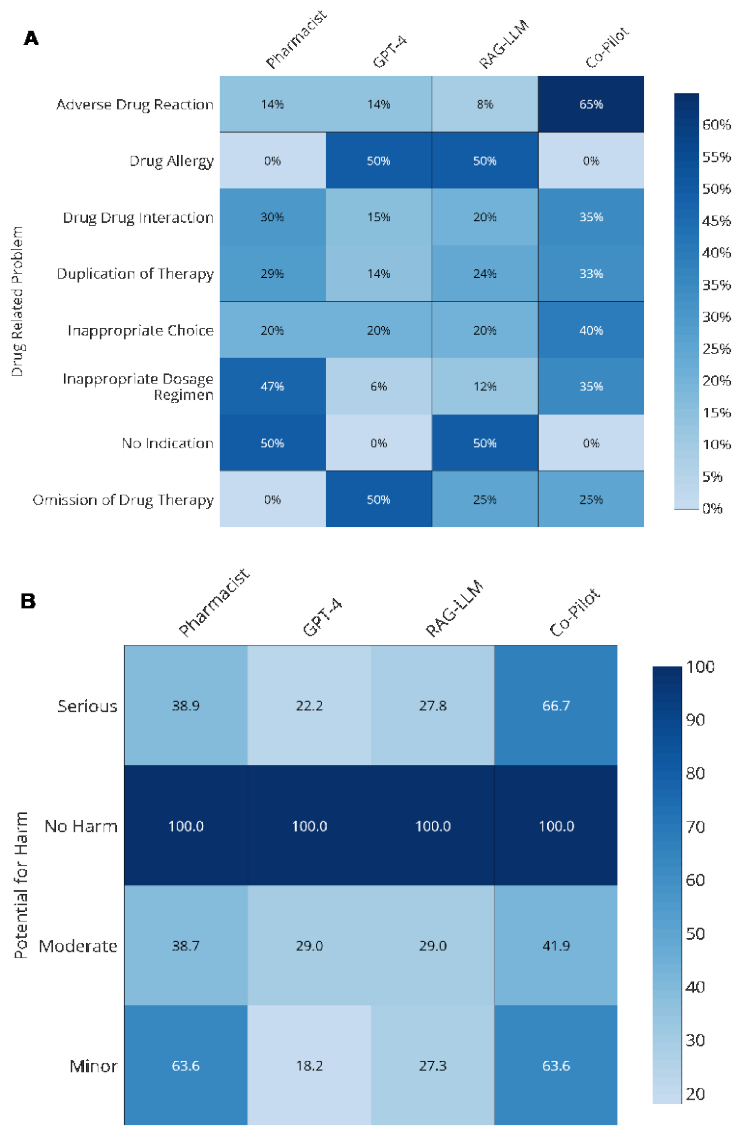

Figure 6: (A) Heatmap showing relative accuracy of different modes in various categories of DRP. (B) Heatmap showing relative accuracy of different modes for DRPs of varying severity of harm.

**Performance of RAG-LLM using Gemini Pro 1.0 and Med-PaLM 2**

We also performed a post-hoc experiment using Gemini-Pro 1.0 and Med-PaLM 2. Both are conducted through Google Cloud Platform, with temperature set as 0.2. We used the same RAG version 2 retrieval, and generated responses using the same prompt and contextual knowledge using Gemini Pro 1.0 and Med-PaLM 2. Results are shown in Table 2.

|  | Gemini Pro 1.0 | | Med-PaLM 2 | | GPT-4 (Version 2) | |
| --- | --- | --- | --- | --- | --- | --- |
|  | **Mean** | **SD** | **Mean** | **SD** | **Mean** | **SD** |
| **Precision** | 0.301 | 0.014 | 0.342 | 0.059 | 0.317 | 0.020 |
| **Recall** | 0.372 | 0.053 | 0.350 | 0.019 | 0.426 | 0.028 |
| **F1** | 0.332 | 0.030 | 0.345 | 0.036 | 0.364 | 0.023 |
| **Accuracy (%)** | 37.200 | 5.300 | 35.000 | 1.900 | 42.600 | 0.280 |

Table 2: Comparative performance of Gemini Pro 1.0 and Med-PaLM 2 based RAG models against GPT-4 based RAG model. GPT-4 based model outperformed Gemini-Pro and Med-PaLM 2 based models in terms of accuracy, Recall and overall F1 score, while Med-PaLM 2 based model demonstrated better precision.

**Discussion**

Although most of recent LLM studies focused heavily on improving model performance based, few studies have explored its function qualitatively and practical applicability of LLM in actual healthcare deployment. In this study, we provide important insights to the feasibility of deploying a novel LLM-based CDSS to healthcare practice. Our selected RAG-LLM model improved the accuracy of DRP identification by 22%. When deployed autonomously, GPT-4 and RAG-LLM showed subpar performance across all metrics of accuracy, precision, recall and F1. However, when deployed as a co-pilot, the RAG-LLM model is capable of highlighting DRPs that were missed by LLM and RAG-LLM. This suggests strongly that for complex reasoning clinical tasks such as medication chart review, LLMs alone are not adequate. A co-pilot mode can potentially improve model utility and also performance of human alone. We also demonstrated comparative performance of 2 different versions of RAG-LLM, illustrating incremental improvement in retrieval from complex medical documents. Hence, it suggested excellent potential of future LLM integration into healthcare practice with ever improving LLM performance. However, we observed that in the co-pilot mode, performance declined in 2 key categories of DRP, namely inappropriate dosage regimen and no indication for drug. This is potentially due to inconsistencies in document formats.[34] For instance, dosage adjustment recommendations in drug monographs may be presented in table format (which poses inherent difficulties for GPT to process), or in a prose format. Indications listed in drug monographs may not be exhaustive. In addition, off-label prescribing practices vary between institutions and not explicitly listed in tertiary drug monographs.

Medication errors often result in unintended patient harm and considerable healthcare cost. As described in a systematic analysis of studies measuring economic impact of medication errors, various cost parameters are involved including hospitalization, medication, primary care, outpatient, litigation and non-healthcare costs. The mean costs per error per study was reported to be as high as €111 727 per error (translating to SGD$172 842 per error).[4] The advent of digital medical records have enabled the proliferation and adoption of CDSS in healthcare systems worldwide. CDSS have been designed for various purposes, mostly customized towards the clinical and operational needs of health institutions and patients.

CDSS can be broadly classified into two categories: knowledge-based and non-knowledge-based systems. Knowledge-based (KB) CDSS rely on a set of rules and associations derived from clinical guidelines or expert opinions. They function by matching patient-specific data with a knowledge base and providing recommendations or alerts.[35] Non-knowledge-based (NKB) systems, on the other hand, use artificial intelligence and machine learning (ML) techniques to analyse large datasets, identify patterns, and make predictions or recommendations based on new data without predefined rules.[7] Our LLM-based CDSS is a novel approach to a NKB based system. Widespread adoption of NKB CDSS systems is met with barriers, including lack of transparency, uncertainty relating to the evidence and lack of trust in the system, or disruptions to the clinical workflow that adds time to routine clinical practice.[36,37] Till date, ML-based CDSS are fairly narrow in applications, most being disease domain specific. LLM grounded with contextual knowledge present with various advantages over ML-based models, including the ability to integrate and process vast amounts of varied data types including unstructured clinical texts, easy updating of clinical knowledge corpus without the need for explicit retraining, offer explanations in natural language that are more comprehensible to human practitioners. We demonstrate that a RAG-LLM CDSS performed equally across different clinical disciplines and medication classes, being agnostic to disease state.

LLMs, especially GPT based models, have been evaluated for their role as decision support tools. One study evaluated the accuracy of GPT-4 base model to generate medication plans using n-gram based evaluation scores like ROUGE.[38] Despite the small sample size and the use of fictional cases, we are the first study to demonstrate the utility of a novel RAG-LLM based CDSS in enhancing safe medication prescribing. We are also the first to report the impact of RAG-based LLM models in improving efficiency and consistency of pharmacists under conditions simulated to reflect real-world workload and patient. In addition, we performed a post-hoc study to provide insights into comparative performance of different state-of-the-art, commercial LLMs in this clinical application. GPT-4 outperformed the other models in most performance metrics.

We were unable to perform a direct comparison of our RAG-LLM model against traditional knowledge-based CDSS. KB CDSS are implemented for various purposes and often tailored to the peculiar needs of the healthcare facility. Regarding their efficacy in reinforcing prescribing safety, various studies have shown that KB CDSS can significantly contribute to the reduction of prescription errors and improve prescribing practices. In a meta-analysis of 68 trials evaluating CDSS on physician prescribing, positive behaviour improved by 4.4% (95% CI 2.6% to 6.2%) with the deployment of KB CDSS.[39] AI-based CDSS have similarly demonstrated positive outcomes on prescribing safety and potential cost savings from DRP avoidance.[40,41] These trials underscore the vital role of CDSS in enhancing prescribing safety, whether through rule-based systems or advanced AI-driven analytics. However, the successful implementation and effectiveness of these systems depend on several factors, including system design, user interface, integration into clinical workflows, and training of healthcare professionals. LLM-based CDSS present with unique advantages such as the ability to process and interpret large volumes of unstructured clinical data, adapt to evolving medical knowledge, and provide personalized treatment recommendations.[42,43] This supports more informed, relevant and individualized patient care decisions.

The limitations of our study are as follows: (1) DRPs identified from clinical scenarios were limited to 5 categories and this might limit generalizability clinical cases with other DRP categories, (2) Small number of fictional clinical scenarios, however the fictional scenarios are highly complex allowing for initial exploration of RAG-LLM based CDSS capabilities, (3) Development and evaluation of RAG-LLM using GPT-4 as sole LLM (4) Bias and fairness of output not evaluated, encoding gender and racial biases has been quantitatively demonstrated for native GPT-4 model[44] (5) Knowledge base from local institution due to copyright issues, (6) Rapid development of new LLM models and versions, limit conclusions from study results, (7) Knowledge source was based on local institutional monographs and drug use guidelines, this may not be broadly applicable across different practice settings, (8) No comparison against current knowledge based or non-knowledge based CDSS.

**Conclusion**

The integration of a RAG-LLM into CDSS presents as a novel tool in improving prescription safety. Simulated with LLM integration into healthcare practice, our study reveals that when used in tandem with junior pharmacists, the RAG-LLM enhances the identification of DRPs, surpassing the accuracy of LLMs alone. The combined pharmacist and LLM model (co-pilot mode) demonstrates superior performance in detecting moderate to severe DRPs and offers a promising hybrid solution for improving patient safety in medication management. This co-pilot model could represent the next step in CDSS development, merging human expertise with the analytical prowess of AI to improve healthcare outcomes.

**Supplement 1: Summary of Case Vignettes**

| Case No | Discipline | Brief Description of Clinical Vignettes | Number of Medications Prescribed | ATC Categories |
|---|---|---|---|---|
| 1 | Cardiology | 61-year-old Malay male with history of diabetes, hypertension, chronic kidney disease, and recurrent vascular thrombotic events. He has undergone multiple surgeries, including amputations and thrombolysis. He presented with chest pain and was diagnosed with an evolved anterior myocardial infarction, for which he underwent successful percutaneous coronary intervention (PCI). | 13 | Electrolytes, Antithrombotics (Heparins), Insulin (fast-acting; rapid-acting and long-acting), Platelet aggregation inhibitors, proton pump inhibitors, Vasodilators, DPP-IV inhibitors, Beta-blocking agent (selective), ACE-inhibitors, B vitamins |
| 2 | Cardiology / Gastroenterology | 41-year-old Malay male with a history of coronary artery disease, diabetes mellitus, KDIGO stage 3 kidney disease, and a recent episode of acute gout. He presented with epigastric pain and was admitted to hospital from the clinic. Investigations included coronary angiography with a view for percutaneous coronary intervention and gastrointestinal scopes. His medication allergies include penicillin. | 13 | Fast-acting insulins, parenteral nutritional products, blood glucose-lowering drugs, oral antidiabetics, lipid-modifying agents, sulfonylureas, beta-blocking agents, agents for gout, macrolides, nitroimidazole derivatives, proton pump inhibitors, antithrombotic agents. |
| 3 | Cardiology | 66-year-old male with a history of minor coronary artery disease, hypertension, hyperlipidaemia, diabetes, and a chronic left caudate nucleus infarct. He presents with left lower limb swelling and pain, fever, and chest tightness. Diagnosed with an evolved inferior STEMI and left lower limb cellulitis complicated by acute kidney injury. | 15 | Long Acting insulin, Heparin group, low molecular weight heparins, Macrolides, Platelet aggregation inhibitors excl. heparin, Angiotensin-converting enzyme inhibitors, Sulfonylureas, Biguanides, Dipeptidyl peptidase 4 (DPP-4) inhibitors, HMG CoA reductase inhibitors, Beta-blocking agents, selective, Dihydropyridine derivatives, Proton pump inhibitors, Organic nitrates, Drugs used in erectile dysfunction |
| 4 | Cardiology | 75-year-old Chinese male with a history of hypertension, hyperlipidaemia, benign prostatic hyperplasia, and a right occipital parasagittal meningioma. He presents with atypical chest pain, which is a dull ache and non-radiating. He has a history of ischemic cardiomyopathy and has undergone coronary artery bypass graft surgery and mitral valve repair. | 6 | Beta-blocking agents, selective, Platelet aggregation inhibitors excl. heparin, 5-alpha-reductase inhibitors, proton pump inhibitors, Alpha-adrenoreceptor antagonists, HMG CoA reductase inhibitors |
| 5 | Cardiology / Respiratory | 58-year-old female admitted for acute pulmonary edema. Her past medical history includes left ataxic hemiparesis, poorly controlled diabetes mellitus, and hypertension. She exhibits shortness of breath, lower limb swelling, and is found to have an NSTEMI with underlying chronic type 2 respiratory failure likely contributed by obstructive sleep apnea. | 16 | Insulins and analogues, Organic nitrates, High-ceiling diuretics, Heparin group, low molecular weight heparins, Angiotensin II antagonists, Beta-blocking agents, selective, Sulfonylureas, Dipeptidyl peptidase 4 (DPP-4) inhibitors, Platelet aggregation inhibitors excl. heparin, Proton pump inhibitors |
| 6 | General Medicine | 58-year-old male with a history of hypertension, admitted with dizziness and vertiginous symptoms, associated with difficulty balancing. He has a strong smoking history and had a possible posterior circulation stroke. The patient also shows signs of hyponatremia, high anion gap metabolic acidosis likely from fasting ketoacidosis, and possible polycythemia related to his smoking history. | 7 | Angiotensin-converting enzyme inhibitors, Platelet aggregation inhibitors excl. heparin, HMG CoA reductase inhibitors, Combination of penicillins, including beta-lactamase inhibitors |
| 7 | Endocrinology | 65-year-old Indian female with a history of type 2 diabetes mellitus, hyperlipidemia, hypertension, osteoarthritis, and a past surgical procedure for a distal radius fracture. She was recently admitted for Group B streptococcus bacteremia secondary to pneumonia and experienced septic shock. She presented with an unwitnessed loss of consciousness and was found to have hypoglycemia and acute kidney injury on admission | 10 | Angiotensin-converting enzyme inhibitors, Other antiepileptics), Sulfonylureas, Biguanides, HMG CoA reductase inhibitors, High-ceiling diuretics, Acetic acid derivatives and related substances |
| 8 | Endocrinology | 68-year-old Chinese female with hypertension and osteoarthritis. She was previously admitted for severe symptomatic hypercalcemia with inappropriately normal parathyroid hormone levels, acute kidney injury, urinary tract infection, hypertensive urgency, anemia, and bilateral lower limb weakness due to cervical myelopathy. She is currently admitted from clinic for hypercalcemia management. | 11 | Calcitonins, Propulsives, Angiotensin II antagonists, Anilides, Antivertigo preparations, Osmotically acting laxatives, Vitamin D and analogues, Dihydropyridine derivatives |
| 9 | Endocrinology / Infectious disease / Vascular | 71-year-old Chinese male with a complex medical history including ischemic heart disease, poorly controlled type 2 diabetes mellitus, peripheral vascular disease, chronic kidney disease stage 4, and moderate-severe dementia. He has been admitted multiple times for fluid overload and is currently admitted for the same issue. The patient also has infected lower limb diabetic ulcers and a painful indwelling catheter. | 10 | Insulins and analogues, High-ceiling diuretics, Biguanides, HMG CoA reductase inhibitors, Platelet aggregation inhibitors excl. heparin, H2-receptor antagonists, Contact laxatives |
| 10 | Ophthalmology | Female patient with right phacomorphic glaucoma and pseudophakia. She has diabetes mellitus, hypertension, hyperlipidemia, and no known drug allergies. There is a history of cataract and glaucoma, with no familial history of glaucoma. | 16 | Carbonic anhydrase inhibitors, HMG CoA reductase inhibitors, Alpha-adrenoreceptor agonists, Angiotensin-converting enzyme inhibitors, Sulfonylureas, Anxiolytics, Prostaglandin analogues, Sodium-glucose co-transporter 2 (SGLT2) inhibitors, Biguanides, Aldosterone antagonists) |
| 11 | Ophthalmology / Endocrinology | 65-year-old female with severe thyroid eye disease (TED) related to Graves' disease. She has undergone total thyroidectomy and bilateral orbital decompression. Her current admission is for postoperative management of TED. | 15 | Combination of penicillins, including beta-lactamase inhibitors, Glucocorticoids, Anilides, HMG CoA reductase inhibitors, Selective COX-2 inhibitors, Thyroid hormones, Proton pump inhibitors, Aminoglycoside antibiotics |
| 12 | Ophthalmology / Infectious Disease | 45-year-old male with no significant past medical history. He presented with right eye endophthalmitis and underwent various treatments, including intravitreal vancomycin and washout procedures. The patient was diagnosed with mycobacterium abscess and was treated with a combination of intravenous and oral medications. | 12 | Carbapenems, Macrolides, Quinolone, Other antibacterials, Anilides, Glucocorticoids, Other opioids |
| 13 | Gastroenterology | 20-year-old female with primary sclerosing cholangitis and Child's Pugh A cirrhosis, chronic ulcerative pancolitis, and a history of cholelithiasis. She presented with left knee pain, fever, and symptoms of a respiratory infection. The preliminary assessment suggests a likely gout flare. | 10 | Immunosuppressants, Bile acids and derivatives, Proton pump inhibitors, Aminosalicylic acid and similar agents, Alpha and beta blocking agents, Quinolone antibacterials, Aminopyrazoles, Imidazole derivatives, Monobactams, Combination of penicillins, including beta-lactamase inhibitors |
| 14 | Gastroenterology | 45-year-old Malay male with a history of Hepatitis C genotype 3A and Child's B cirrhosis. He was admitted for deranged liver function tests and presented with jaundice. He presented with a needle stick injury and agrees to start treatment with Epclusa for Hepatitis C. Other active issues also includes managing possible acute cholecystitis and fluid restriction | 6 | High-ceiling diuretics, proton pump inhibitors, Aldosterone antagonists, Anilides<br>*Epclusa not listed |
| 15 | Gastroenterology | 64-year-old Indian male with a history of Child's B8 Cirrhosis, diabetes, hypertension, hyperlipidemia, ischemic heart disease, atrial fibrillation, stage 3 chronic kidney disease, and hypothyroidism. He presents with worsening abdominal distension, lower limb edema, and reduced urine output. The patient has a history of spontaneous bacterial peritonitis and is on lifelong Ciprofloxacin for SBP prophylaxis. | 10 | Organic nitrates, Dipeptidyl peptidase 4 (DPP-4) inhibitors, Aminopyrazoles, Proton pump inhibitors, Platelet aggregation inhibitors excl. heparin, Alpha and beta blocking agents, Quinolone antibacterials, Cardiac glycosides, Thyroid hormones |
| 16 | Gastroenterology | 77-year-old female with a history of Child's C10 cryptogenic liver cirrhosis with portal vein hypertension. She presented with diuretic- | 8 | Glycopeptide antibacterials, Antibiotics, Imidazole derivatives, Proton pump inhibitors, |

| | | resistant ascites s/p ascitic drain with H. pylori infection. | | Tetracyclines, Osmotically acting laxatives, Quinolone antibacterials, Dihydropyridine derivatives) |
|---|---|---|---|---|
| 17 | General Surgery / Oncology | 55-year-old Chinese male with a history of perforated duodenal ulcer, epithelial carcinoma of the salivary gland with liver metastasis, and neutropenia sepsis. His current condition includes postoperative care following laparoscopic omental patch repair of a perforated duodenal ulcer and chemotherapy for his cancer. | 10 | Insulins and analogues, Echinocandins, Colony-stimulating factors, Proton pump inhibitors, Anilides, Combination of penicillins, including beta-lactamase inhibitors, Other opioids, Glycopeptide antibacterials, Nucleoside and nucleotide reverse transcriptase inhibitors |
| 18 | Vascular Surgery | 71-year-old Malay female with a complex medical history including end stage renal failure post living donor kidney transplant, complicated by chronic kidney disease of allograft, left renal cell carcinoma post radical nephrectomy, hyperparathyroidism with hypercalcemia, hypertension, type 2 diabetes mellitus, benign prostatic hyperplasia, and high cholesterol. She presents with right big toe pain and duskiness, leading to a diagnosis of dry gangrene of the right first toe. | 18 | Insulins and analogues, Beta-blocking agents, selective, HMG CoA reductase inhibitors, Immunosuppressants, Other calcimimetics, Platelet aggregation inhibitors excl. heparin, Trimethoprim and derivatives, combinations with sulfamethoxazole, Other lipid modifying agents, Folic acid, High-ceiling diuretics, Sulfonylureas, Dipeptidyl peptidase 4 (DPP-4) inhibitors, Angiotensin-converting enzyme inhibitors, Magnesium, Mycophenolic acid, proton pump inhibitors, Alpha-adrenoreceptor antagonists) |
| 19 | General Surgery / Colorectal Surgery | 60-year-old Chinese male with a history of diabetes, hypertension, hyperlipidemia, polycythemia rubra vera, end-stage renal failure on hemodialysis, diverticular disease, and past surgery for perforated jejunal diverticulitis. He presented with lower abdominal pain and constipation suspicious for intestinal obstruction and later developed anal pain. | 17 | Osmotically acting laxatives, Anilides, Contact laxatives, Local anesthetics, Other opioids, HMG CoA reductase inhibitors, Other antiepileptics, Dipeptidyl peptidase 4 (DPP-4) inhibitors, Sulfonylureas, Propulsives, Other calcimimetics, Vasodilators used in peripheral vascular diseases, Phosphate binders, Organic nitrates, Quinolone antibacterials |
| 20 | General Surgery | 46-year-old male with a history of adjustment disorder, lupus nephritis, hypertensive urgency, and chest pain. He presented with right lower abdominal pain associated with fever, diagnosed as perforated appendicitis. He underwent a laparotomy converted to open limited right hemicolectomy. | 12 | Potassium, Third-generation cephalosporins, Imidazole derivatives, Proton pump inhibitors, Serotonin (5HT3) antagonists, Propulsives, Anilides, Dihydropyridine derivatives, Angiotensin-converting enzyme inhibitors, Non-steroidal anti-inflammatory and antirheumatic products, coxibs, Hydralazine and diuretics, Selective COX-2 inhibitors |
| 21 | General Surgery | 71-year-old male with a diagnosis of cecal diverticulitis and a small diverticular abscess. His past medical history includes benign prostatic hyperplasia, hypertension, and hyperlipidemia. The patient presented with right iliac fossa pain and had a history of abdominal pain. | 11 | Potassium, Third-generation cephalosporins, Imidazole derivatives, Proton pump inhibitors, Propulsives, 5-alpha-reductase inhibitors, Alpha-adrenoreceptor antagonists, Anilides, Other opioids, Sulfonamides, plain, Phenothiazines with aliphatic side-chain |
| 22 | Urology | 75-year-old male with a history of ischemic colitis, hemorrhoids, benign prostatic hyperplasia (BPH), asthma, hypertension, gastritis, ischemic cardiomyopathy, and prostate cancer. He was admitted for gross hematuria and underwent bladder cystoscopy and cystodiathermy. His current issues include gross hematuria likely from prostate cancer and a urinary tract infection | 14 | Potassium, Combination of penicillins, including beta-lactamase inhibitors, Other antineoplastic agents, Platelet aggregation inhibitors excl. heparin, Beta-blocking agents, selective, 5-alpha-reductase inhibitors, Anilides, Alpha-adrenoreceptor antagonists, Acetic |

| | | | | acid derivatives and related substances, Organic nitrates, High-ceiling diuretics, Osmotically acting laxatives, HMG CoA reductase inhibitors |
|---|---|---|---|---|
| 23 | Urology | 67-year-old male with a history of hypertension, type 2 diabetes mellitus, ischemic heart disease, benign prostatic hyperplasia, and pyelonephritis. He was admitted for pyelonephritis with persistent purulent discharge from the left loin and underwent percutaneous drainage. | 12 | Insulins and analogues, Platelet aggregation inhibitors excl. heparin, Beta-blocking agents, selective, Combination of penicillins, including beta-lactamase inhibitors, Other antiepileptics, Osmotically acting laxatives, Anilides, Other opioids, Alpha-adrenoreceptor antagonists, |

**Supplement 2: DRPs and Risk / Potential for Harm Categories**

| Case No | DRP(s) Category and Description | Severity / Potential for Harm |
|---|---|---|
| 1 | (1) Drug allergy: Background of NSAIDS allergy (rash) but prescribed with aspirin without any challenge or test dose | Moderate |
| | (2) Inappropriate dosage regimen: Enoxaparin dosed at 60mg BD in obese patient of 90kg, no bleeding. | Serious |
| | (3) Omission of therapy: Omission of statin therapy in patient presenting with myocardial infarction | Moderate |
| | (4) Adverse drug reaction: Borderline blood pressure but prescribed with perindopril and bisoprolol | Moderate |
| 2 | (1) Drug Drug Interaction: Significant Interaction between atorvastatin and clarithromycin | Moderate |
| | (2) Inappropriate dosage regimen: Colchicine dosed in MG instead of MCG | Serious |
| | (3) Adverse drug reaction: Bradycardia but prescribed with both bisoprolol and ticagrelor | Moderate |
| 3 | (1) Wrong indication: Wrong drug of clarithromycin instead of clindamycin for cellulitis in patient with penicillin allergy | Serious |
| | (2) Inappropriate dosage regimen: Wrong dose of enoxaparin (dosed 1mg/kg BD) in patient with renal failure | Moderate |
| | (3) Drug drug interaction: Significant drug interaction (contraindicated) between sildenafil and isosorbide mononitrate | Serious |
| 4 | Control Case | NA |
| 5 | (1) Adverse drug reaction: Patient presented with hypokalemia, on intensive furosemide therapy without electrolyte replacement | Serious |
| | (2) Adverse drug reaction: Patient presenting with acute pulmonary edema but continued on beta-blocker therapy | Moderate |
| | (3) No indication for medication: Patient on triple antithrombotic therapy with aspirin / clopidogrel / enoxaparin post myocardial infarction, also continued on dipyridamole (chronic medication for previous history of CVA) | Serious |
| 6 | (1) Duplication of therapy: Duplication between Simvastatin and Atorvastatin. Both belong to the same category of HMG-CoA reductase inhibitor | Minor |
| | (2) Inappropriate dosing frequency for atorvastatin. It is usually given once daily, but does not exceed max daily dose | Minor |
| | (3) Adverse Drug Reaction: Patient has allergy to amoxicillin (rash) but was prescribed with Co-amoxiclav which contains amoxicillin | Moderate |
| 7 | (4) Adverse Drug Reaction: Acute kidney injury with hyperkalemia but continued on enalapril | Moderate |
| | (5) Adverse drug reaction: Acute kidney injury but prescribed with NSAID | Moderate |
| | (6) Adverse drug reaction: Presented with hypoglycemia but continued on glipizide | Serious |
| 8 | (1) Inappropriate dosage regimen: Wrong weight-based dose of calcitonin prescribed for patient with hypercalcemia of malignancy | Serious |
| | (2) Adverse drug reaction: Colecalciferol not held off despite hypercalcemia | Moderate |
| | (3) No indication: Wrong drug of cinnarizine prescribed instead of cinacalcet for patient with b/g secondary hyperparathyroidism | Minor |
| 9 | (1) Duplication of therapy: Patient with poorly controlled diabetes on both ultra-short-acting insulin Aspart and short-acting insulin Actrapid. | Moderate |
| | (2) Omission of therapy: Omisson of basal insulin such as insulatard in patient with poorly controlled diabetes | Moderate |
| | (3) Omission of therapy: Antibiotics not ordered for patient with suspected sepsis | Serious |
| 10 | (1) Adverse drug reaction: Laboratory results demonstrating non-anion gap metabolic acidosis with respiratory compensation, continued on acetazolamide | Moderate |
| | (2) Adverse drug reaction: Hydroxyzine (first generation antihistamine with significant anticholinergic effects) contraindicated in acute-angled glaucoma | Moderate |
| | (3) Adverse drug reaction: Patient with frequent urinary tract infection but continued on dapaglifozin, an SGLT-2 inhibitor | Moderate |
| 11 | (1) Duplication of therapy: Patient prescribed with PO prednisolone and IV Methylprednisolone (high-intensity) concurrently | Minor |
| | (2) Inappropriate dosage regimen: Levothyroxine prescribed in MG instead of MCG | Serious |
| 12 | (1) Drug drug interaction: Significant interaction between tramadol and linezolid leading to increased risk for serotonin syndrome | Moderate |
| | (2) Duplication of therapy: Patient prescribed with IV Tienam and PO ciprofloxacin for gram-negative coverage of urinary tract infection | Minor |
| | (3) Inappropriate dosage regimen: Instructions for prednisolone eye drops was to administer to left eye, when the affected eye was the right | Serious |
| 13 | (1) Drug drug interaction: Significant (contraindicated) interaction between azathioprine and allopurinol leading to increased risk for neutropenia | Serious |
| | (2) Omission of therapy: Patient initiated on allopurinol but not on anti-inflammation therapy to avoid exacerbation of gout flare (e.g. colchicine or steroids) | Moderate |
| 14 | (1) Omission of therapy: Noted complains of no bowel movement for n-days but no laxatives ordered for patient with a significant history of liver cirrhosis | Moderate |
|  | (2) Drug drug interaction: Significant interaction between Epclusa and omeprazole that will reduce absorption and efficacy of anti-viral agent | Moderate |
|  | (3) Inappropriate dosage regimen: Dosage of paracetamol not adjusted in the presence of liver cirrhosis | Moderate |
| 15 | Control Case | NA |
| 16 | (1) Drug drug interaction: Significant interaction between calcium supplement and tetracycline. Calcium reduces absorption of tetracycline when taken concurrently | Moderate |
|  | (2) Inappropriate dosage regimen: Dose of metronidazole not adjusted in the presence of significant liver impairment | Moderate |
| 17 | (1) Inappropriate dosage regiment: The infusion rate of vancomycin has exceeded the maximum recommended rate, leading to increased risk for red man's syndrome | Serious |
|  | (2) No indication: G-CSF (filgrastim) continued despite recovery of absolute neutrophil counts | Moderate |
| 18 | (1) Drug drug interaction: Significant interaction between ciclosporin and atorvastatin, with max dose of atorvastatin limited to not more than 20mg per day | Moderate |
|  | (2) Inappropriate dosage regimen: Inappropriate dose of co-trimoxazole, prescribed in trimethoprim component instead of co-trimoxazole for PCP prophylaxis | Moderate |
| 19 | (1) Inappropriate dosage regimen: Dose of tramadol not adjusted in the presence of significant renal failure | Moderate |
|  | (1) Inappropriate dosage regimen: Dose of gabapentin not adjusted in the presence of significant renal failure | Moderate |
|  | (2) Omission of therapy: Aspirin omitted in patient with significant cardiac history and low risk of bleeding | Minor |
|  | (3) Drug drug interaction: significant interaction between sulphonylurea and ciprofloxacin leading to increased risk for hypoglycemia | Minor |
| 20 | (1) Duplication of therapy: Celecoxib and etoricoxib overlapping mechanism of action | Minor |
|  | (2) Duplication of therapy: PO and IV omeprazole both ordered | No harm |
|  | (3) Adverse drug reaction: Initiation of hydralazine in patient with a background of lupus disease | Serious |
| 21 | (1) Adverse Drug Reaction: Hyperkalemia but continued on potassium chloride infusion | Serious |
|  | (2) Inappropriate dosage regimen: Exceeded maximum recommended dose of omeprazole for the indication of ulcer prevention | Minor |
|  | (3) Drug drug interaction: Significant interaction between prochlorperazine and metoclopramide | Serious |
|  | (4) Omission of therapy: Untreated hyperlipidemia | Minor |
|  | (5) Omission of therapy: Untreated hyperkalemia | Serious |
| 22 | (1) Inappropriate dosage regimen: Rapid intravenous infusion of potassium chloride 10mmol over 1 minute | Moderate |
|  | (2) Omission of therapy: Omission of steroid (e.g. prednisolone) when on abiraterone treatment for castrate resistant bladder cancer | Moderate |
|  | (3) Adverse drug reaction: Diclofenac use in patient with recent myocardial infarction presenting with heightened risk for cardiovascular events | Serious |
| **23** | (1) Duplication of therapy: Prescription of 2 alpha-blockers (Tamsulosin and Alfuzosin) in a patient with history of benign prostate hyperplasia | Minor |
|  | (2) Wrong choice of therapy: Co-amoxiclav use in patient with wound culture growing *E.Coli* reported to be resistant to co-amoxiclav | Moderate |